\documentclass{article}
\usepackage{amsmath}
\usepackage{amssymb}
\usepackage{amsthm}
\usepackage{bm}
\usepackage{graphicx}
\usepackage{multirow}
\usepackage{subcaption}
\usepackage{booktabs}
\usepackage{svg}
\usepackage{xfrac}
\usepackage{titling}
\usepackage[final]{corl_2020} 
\usepackage{authblk}

\usepackage{mathtools}

\definecolor{lidarcolor}{HTML}{4682B4}
\definecolor{radarcolor}{HTML}{FF6347}
\definecolor{predcolor}{HTML}{DAA520}
\definecolor{xdcolor}{HTML}{1D12E3}

\newcommand{\lidarcolortext}[1]{\textcolor{lidarcolor}{#1}}
\newcommand{\radarcolortext}[1]{\textcolor{radarcolor}{#1}}
\newcommand{\predcolortext}[1]{\textcolor{predcolor}{#1}}
\newcommand{\xdcolortext}[1]{\textcolor{xdcolor}{#1}}

\newcommand{\networkname}[0]{LiRaNet}
\newcommand{\datasetname}[0]{X17k}

\newcommand{\tablestyle}[2]{\setlength{\tabcolsep}{#1}\renewcommand{\arraystretch}{#2}\centering\footnotesize}
 \newlength\savewidth\newcommand\shline{\noalign{\global\savewidth\arrayrulewidth\global\arrayrulewidth 1pt}\hline\noalign{\global\arrayrulewidth\savewidth}}

\title{\networkname: End-to-End Trajectory Prediction using Spatio-Temporal Radar Fusion}

\author{Meet Shah*\textsuperscript{1}}
\author{Zhiling Huang*\textsuperscript{1}}
\author{Ankit Laddha*\textsuperscript{1}}
\author{Matthew Langford\textsuperscript{1}}
\author{Blake Barber\textsuperscript{1}}
\author{Sidney Zhang\textsuperscript{1}}
\author{Carlos Vallespi-Gonzalez\textsuperscript{1}}
\author{Raquel Urtasun\textsuperscript{1,2}}

\affil{\textsuperscript{1}Uber Advanced Technologies Group, \textsuperscript{2} University of Toronto 
\authorcr{
    \tt\small \{meet.shah, zhiling, aladdha, mlangford, \\ 
    \tt\small bbarber, sidney, cvallespi, urtasun\}@uber.com}
    \vspace{-2em}
}

\begin{document}
\maketitle

\thispagestyle{empty}
\makeatletter{\renewcommand*{\@makefnmark}{}
\footnotetext{*Equal Contribution}\makeatother}

\begin{abstract}
In this paper, we present {\networkname}, a novel end-to-end trajectory prediction method which utilizes radar sensor information along with widely used lidar and high definition (HD) maps. Automotive radar provides rich, complementary information, allowing for longer range vehicle detection as well as instantaneous radial velocity measurements.
However, there are factors that make the fusion of lidar and radar information challenging, such as the relatively low angular resolution of radar measurements, their sparsity and the lack of exact time synchronization with lidar. To overcome these challenges, we propose an efficient spatio-temporal radar feature extraction scheme which achieves state-of-the-art performance on multiple large-scale datasets.
Further, by incorporating radar information, we show a $52\%$ reduction in prediction error for objects with high acceleration and a $16\%$ reduction in prediction error for objects at longer range.
\end{abstract}


\section{Introduction}
\vspace{-0.8em}

Trajectory prediction plays a pivotal role in the success of self-driving vehicles (SDVs) in dynamic and interactive environments. The SDV can plan a safe path by taking into account the future position of objects to preemptively avoid potentially harmful interactions. It involves understanding each actor's motion and intention as well as the complex interactions with other actors and the environment. Due to its importance for SDVs, a plethora of work in trajectory prediction has been undertaken~\cite{dp,intentnet,tensorfusion,spagnn,multixnet,multipath}.

Recently, end-to-end approaches that directly predict trajectories from raw sensor data have gained traction~\cite{faf, intentnet, stinet, spagnn, multixnet, laserflow, rvfusenet} due to their efficiency and high performance. These approaches typically exploit lidar as well as HD maps for this task. Lidar captures positional information about objects, while HD maps provide scene context and a prior on possible paths. To accurately predict future motion, these methods need to implicitly understand the dynamics of objects in the scene. They do so by relying on a sequence of position observations from lidar over time, increasing the reaction time of the vehicle. This can be problematic in certain critical situations where accurate, low-latency motion prediction is needed to avoid an imminent collision. 

\begin{figure}[ht]
    \centering
    \includegraphics[width=\textwidth]{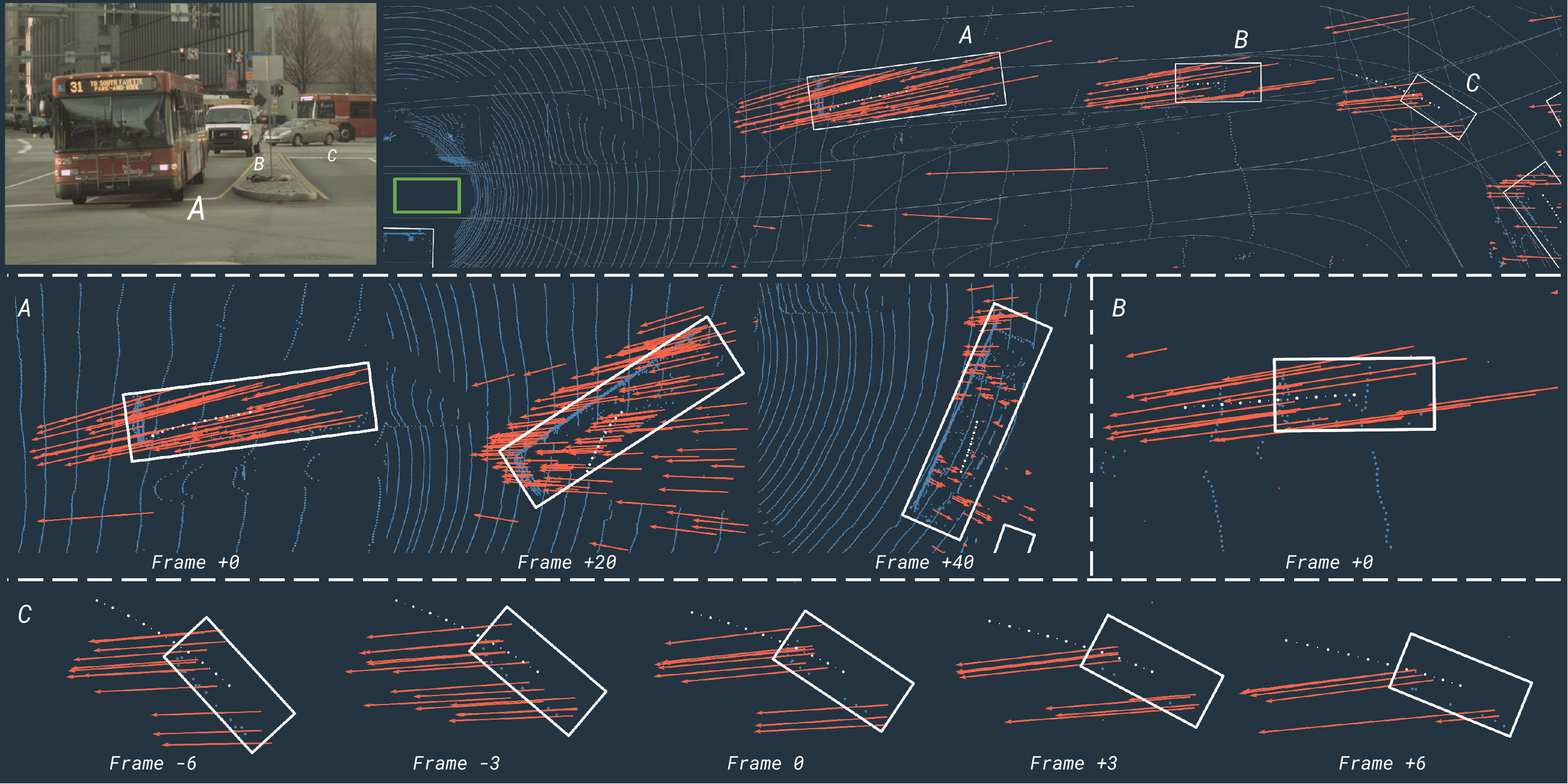}
    \caption{
        \textbf{Challenges and Benefits of Radar:} An example scene from {\datasetname} in bird's eye view where \lidarcolortext{lidar points (light blue)} and \radarcolortext{radar point velocities (orange)} are visualized with labels (white) for current, past and future frames.
        Vehicle A is a turning bus that has multiple radar points across frames. By effectively combining them over space and time a full 2D velocity and turning rate can be recovered. 
        Vehicle B shows the high positional noise that inherently comes with radar.
        Vehicle C shows a case with sparse lidar points where implicitly associating them across time can be challenging. However, radar points present around C can add context for the model to detect and predict the trajectory.
        More examples (with model predictions) where radar data can positively complement lidar can be found in Fig~\ref{fig:qualitative_improvements}.
    \vspace{-1.8em}
    }
    \label{fig:overview}
\end{figure}

Radar, on the other hand, provides instantaneous velocity and positional information. Radar has been deployed in numerous applications in the past half century ranging from subterranean to space\footnote{\url{https://en.wikipedia.org/wiki/radar}} and more recently, has been deployed as a primary sensing modality in commercial ADAS systems\footnote{\url{https://en.wikipedia.org/wiki/Advanced_driver-assistance_systems}}. Given the ubiquity of radar sensing in many domains for trajectory prediction, it is surprising that it has not featured more prominently in end-to-end prediction systems for self-driving. However, the benefit of radar in terms of instantaneous velocity measurement also comes with challenges. Typical automotive radars have lower resolution and provide point clouds with higher position uncertainty (Fig.~\ref{fig:overview}) than typical lidars. Furthermore, it only provides the radial component of the object velocity as depicted in Figure~\ref{fig:overview}. As a result, most existing approaches~\cite{radarnet,cho2014multi,gohring2011radar} utilize radar by heavily relying on late fusion. They leverage the object's position for associating radar points to the object, and the radial velocity measurement from these associated radar points to estimate object velocity. Therefore, they require a multi-stage (detect-track-predict) method to utilize a temporal sequence of radar and for predicting trajectories. 

In contrast, we directly do an early fusion of a temporal sequence of radar data along with lidar and map for trajectory prediction. We propose a novel end-to-end approach, called {\networkname}, to fuse dynamic information from radar with lidar and HD maps. In particular, we present an \textit{early fusion} method that effectively combines radar data with other sensors at feature level. To overcome the sparsity, positional uncertainty and partial velocity measurements of radar, our method fuses a sequence of past radar data to learn spatio-temporal features on a bird's eye view (BEV) grid using a graph. These dynamics-rich radar based features are combined with lidar and map features containing detailed geometric information to accurately predict the future positions of objects.

Our proposed method provides state-of-art trajectory prediction performance on two large-scale datasets. We demonstrate that by effectively exploiting radar information our approach provides substantial improvements on three different trajectory prediction formulations. Furthermore, we showcase that adding radar significantly improves trajectory prediction in safety and latency critical scenarios where lidar-only methods struggle due to sparse points and rapidly changing dynamics.

\section{Related Work}
\vspace{-0.8em}

\textbf{Trajectory Prediction:} Classical methods~\cite{dp, rulesofroad, tensorfusion, desire, precog, mfp, multipath} utilize a multi-stage detect-track-predict paradigm for trajectory prediction. In the first stage, sensor data is used to detect objects~\cite{pixor, lasernet,lasernet++, pointpillars, pointrcnn, mvf, std} at each time step independently. These objects are then tracked across time to generate a temporal sequence of object positions. The sequence of positions are then used for trajectory prediction. By dividing the problem into stages, independent progress can be made for each step and different sensors can be fused at different stages. However, such approaches suffer from cascading errors and high latency since multiple models are run sequentially. This motivates the recent usage of end-to-end single stage methods that can potentially improve performance.

The seminal work of~\cite{faf} proposed a single stage end-to-end approach for predicting trajectories directly from raw lidar data using a BEV grid. This approach is further improved in subsequent works by incorporating HD map~\cite{intentnet}, reasoning about interactions~\cite{spagnn,stinet}, using a better network architecture~\cite{multixnet,stinet} and range view based lidar representation~\cite{laserflow,rvfusenet}. As compared to multi-stage approaches, these methods are faster, easier to maintain in a production environment and have higher performance~\cite{spagnn}. They can easily take advantage of sensor information and context of other objects for predictions. In this paper, we propose to add radar along with lidar and HD maps for improving end-to-end trajectory prediction.

\textbf{Multi-Hypothesis Trajectory Prediction:}
Many single stage methods produce a single deterministic trajectory~\cite{intentnet,stinet} or a single distribution such as Gaussian~\cite{spagnn} or Laplacian~\cite{laserflow,rvfusenet,multixnet} for each waypoint of the trajectory. However, to ensure safe and efficient operations, an autonomous vehicle is required to anticipate a multitude of possible behaviors of actors in the scene. This has motivated a plethora of work~\cite{mtp, multipath,covernet,mfp,socialgan,desire} in probabilistic multi-hypothesis trajectory prediction. We incorporate the mixture distribution formulation of ~\cite{mtp} and~\cite{multipath} in our proposed method and demonstrate improvements by incorporating radar.

\textbf{Radar for Self Driving:} Automotive radars have several desirable properties such as cost effectiveness and resilience to extreme weather. These have led to it being a widely used sensor for ADAS and has recently motivated diverse research efforts to utilize radar information for self driving. However, most existing methods utilize radar only for improving perception~\cite{chavez2015multiple, gohring2011radar, cho2014multi, chadwick2019distant, meyer2019deep, major2019vehicle, radarnet} which has an indirect impact on trajectory prediction through a multi-stage pipeline. In comparison, our proposed approach directly improves trajectory prediction. Methods utilizing radar data can be divided into two major groups: tracking based and learning based. 

Classical tracking based approaches~\cite{chavez2015multiple, gohring2011radar, cho2014multi} fuse radar data with other sensors using filtering techniques such as Kalman Filters to create tracks. Dynamics information in each track can be integrated forward for trajectory prediction. However, data driven approaches based on deep learning can easily outperform~\cite{covernet, multipath} such simple prediction methods. Recently, several learning based approaches using radar data for self driving have been proposed. UNets are used in~\cite{sless2019road, lombacher2017semantic} to learn semantic segmentation of BEV grid using only radar data. Radar and Camera are fused for 3D object detection in~\cite{chadwick2019distant,meyer2019deep,nobis2020deep}. Nonetheless, rich velocity information provided by radar is ignored in all of these approaches. More recently, joint 3D detection and velocity estimation is performed in~\cite{major2019vehicle} using raw Range-Azimuth-Doppler tensor and in~\cite{radarnet} by fusing lidar and radar. In contrast to these, we propose to utilize radar for trajectory prediction in an end-to-end learning based method. 

\section{Multi-Sensor Trajectory Prediction}
\vspace{-0.8em}

\begin{figure}[ht]
    \centering
    \includegraphics[width=\textwidth]{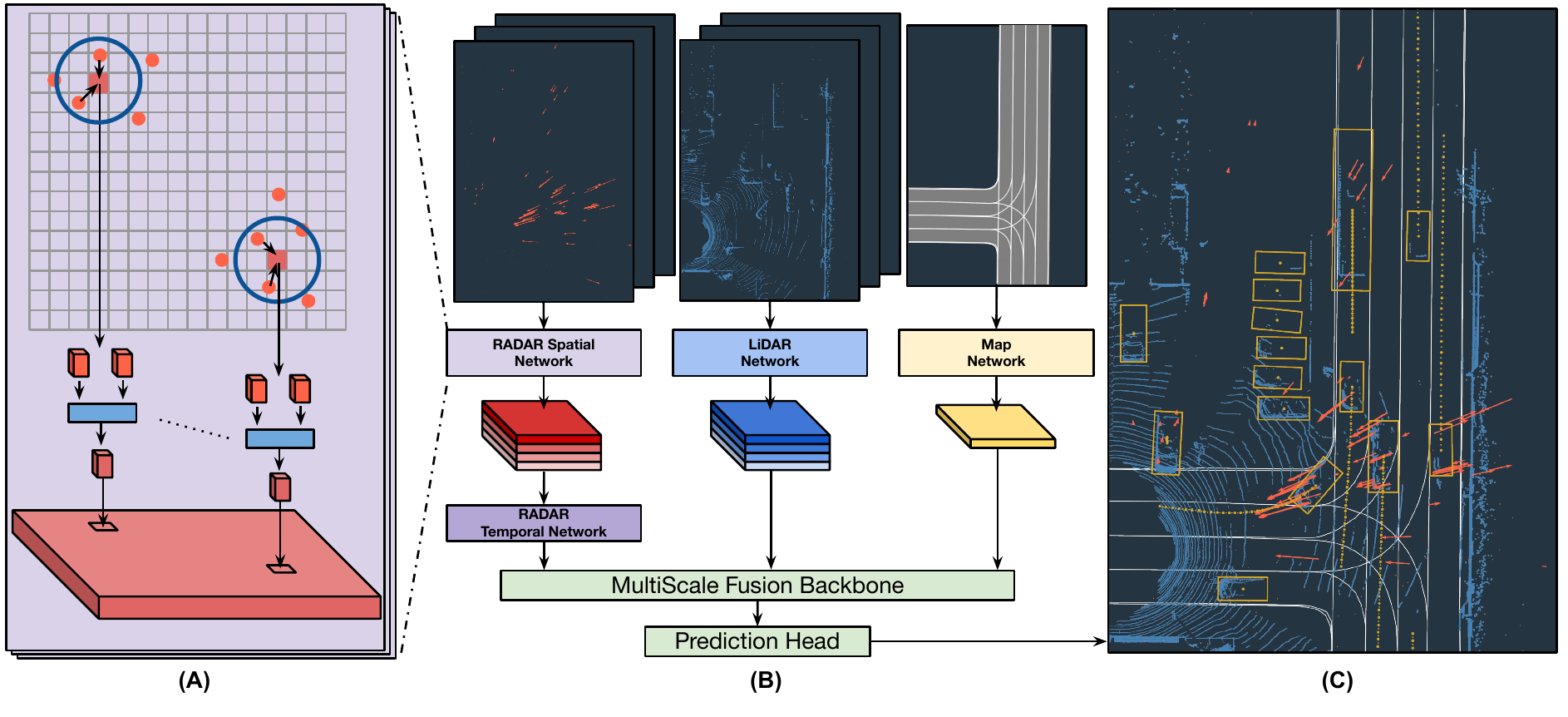}
    \caption{
    \textbf{{\networkname} overview:}
    %
    %
    Our radar feature extraction network \textbf{(A)} extracts spatio-temporal features from raw radar points in 2 steps: 
    (1) for each frame we create a graph between the BEV grid cells and radar points to learn spatial features of each cell using a non-rigid convolution, 
    (2) these spatial features are further fused temporally by stacking across channel dimension and using an MLP to get a radar feature volume.
    This feature volume is then fused with feature volumes from other sensors and fed to a joint perception-prediction network \textbf{(B)} which produces detections and their future trajectories. An example prediction for a scene from {\datasetname} can be seen in \textbf{(C)}.
    }
    \vspace{-1.2em}
    \label{fig:model}
\end{figure}

We propose a novel approach for end-to-end trajectory prediction by fusing velocity-rich information from radar with other sensor data. We use a Bird's Eye View (BEV) grid which is commonly used to fuse multiple sensors~\cite{ibm} as well as temporal sequences of 3D sensors~\cite{faf}. Thus, we propose to learn spatio-temporal features for BEV grid cells using a sequence of past radar data. Figure ~\ref{fig:model} shows an overview our proposed approach. Each input sensor modality is first processed individually to generate multiscale BEV feature volumes. Feature volumes from multiple modalities are then merged together using concatenation and then passed through a backbone network to generate features for each cell in the grid. These are then fed into our object detection and trajectory prediction headers to obtain predictions. 

In the following sections, we first describe data from radar sensors and the challenges associated with it. Next, we describe our proposed approach to learn features for BEV grid cells using radar. Then, we describe other sensor representations and network architecture used to fuse all the modalities. Finally, we discuss the end-to-end training scheme of our proposed network.

\subsection{Radar Sensor Data}
\label{sec:radar_overview}
\vspace{-0.8em}
Radar observes the environment by transmitting and receiving radio waves. The raw data cube generated by the radar is generally too large to be practically relayed off-sensor or processed without specialized hardware. Off the shelf automotive radar sensors overcome this by detecting peaks in the raw radar data cube using a Constant False Alarm Rate (CFAR)~\cite{richards2010principles} algorithm, and only reporting these peaks (the rest of the data is assumed to be noise). Given this processing, we consider these peaks, called radar points, for feature learning.

Until now, radar data has not been used for end-to-end trajectory prediction in self driving due to a few key difficulties and shortcomings. 
Current-generation automotive radar sensors can: contain 2-3 orders of magnitude fewer points relative to lidar, have angular accuracies 1-2 orders of magnitude worse relative to lidar, and while they do provide a direct observation of velocity, they can only observe velocity in the radial direction. All of these challenges complicate the fusion of lidar and radar. However, intuitively the complementary radial velocity measurement of radar should be able to help in improving the estimate of object dynamics. In the following section we propose a method to overcome these challenges to  effectively learn features for BEV grid cells and improve prediction.

\subsection{Spatio-Temporal Feature Learning Using Radar}
\vspace{-0.8em}
Multiple radar sensors are usually placed on a vehicle to achieve a $360^{\circ}$ view. 
We create a radar sweep such that it contains the measurements from all radars in a time interval. We assume we are given a sequence of $M$ such radar sweeps as input. All radar points are transformed into the coordinate frame associated to the most recent sweep. 
For each radar point $i$ in the sweep $m$, we calculate a feature vector $\bm{f}_{i}^{m}$ containing its 2D position ($x_{i}^{m}, y_{i}^{m}$), radar cross-section (RCS) ($e_{i}^{m}$) and ego-motion compensated radial velocity (expressed as a 2D vector in the vehicle coordinate frame) ($v_{x,i}^{m}, v_{y,i}^{m}$). 
We propose to use two separate modules to learn features from radar points for BEV cells: spatial and temporal. The spatial module works on a each  sweep individually and provides robustness to sparsity and position errors observed in radar points. The temporal learning module provides robustness to radial velocity observation by using a sequence of sweeps to implicitly recover full 2D velocity of objects.

\textit{Spatial Feature Learning:} The goal of this module is to extract BEV features for each cell in the BEV grid using a single sweep of radar points. The grid is chosen to be centered at the ego-vehicle position at the most recent sweep. In comparison to lidar, radar is very sparse and has high positional noise. Thus, directly discretizing the radar x,y coordiantes at the same resolution as lidar for feature extraction does not work well (see Sec~\ref{sec:result}). Therefore, we propose the following novel feature learning method.

We first select an appropriate resolution for feature learning. This resolution is usually lower than the resolution for lidar due to sparsity in radar points. We then use a learnable operator to extract the features at this resolution. Unfortunately, we cannot use the standard convolutional operators since the source domain (unordered points) is not a grid like structure. Therefore, we use the recently proposed parametric continuous convolutions~\cite{parametric} for computing the BEV cell features. 
Parametric convolution generalizes the standard convolution operator to non-grid like structure and can handle different input and output domains with pre-defined correspondence between them. In particular we use the more expressive adaptation of~\cite{ibm}.

In our case, the input domain consists of radar points and output domain consists of BEV cells. 
Therefore, for each cell $j$ we can calculate the features $\bm{h}_{j}^{m}$ for the sweep $m$ as (see Figure~\ref{fig:model}):
\begin{equation}
    \bm{h}_{j}^{m} = \sum_{i \in \bm{A}_{j}^{m}} g^{m}\left(\bm{f}_{i}^{m} \oplus (\bm{x}_{i}^{m} - \bm{x}_{j}^{m})\right)
\end{equation}

where $\bm{A}_{j}^{m}$ is the set of associated radar points, $\bm{x}_{i}^{m}$ is the 2D coordinates of the associated radar point, $\bm{x}_{j}^{m}$ is the 2D coordinate of the BEV cell's center, $\oplus$ denotes the concatenation operation, $\bm{f}_{i}^{m}$ is the feature vector for the radar point and $g^{m}(\cdot)$ is an multi-layer perceptron (MLP) with learnable weights shared across all the cells. We calculate $\bm{A}_{j}^{m}$, using nearest neighbor algorithm with a distance threshold. By using a threshold larger than the size of a cell, our method compensates for positional errors in radar. 

\textit{Temporal Feature Learning:} The goal of this module is to combine the spatial features for all the sweeps. For each cell $j$, we calculate the final spatio-temporal feature vector ($h_{j}$) by concatenating the per sweep features $h_{j,m}$ and using an MLP to combine them. 

\subsection{Lidar and HD Map Feature Learning}
\vspace{-0.8em}
We assume that we are given $L$ past lidar sweeps and an HD map of the surroundings. We represent the HD map~\cite{intentnet,multixnet} as a rasterized BEV image containing geometric lane information. We employ a lightweight network to learn map features for BEV cells using the rasterized image.
We represent lidar as an BEV occupancy grid~\cite{faf}. Since lidar is dense and has very low position noise, we can assume that the lidar point coordinates ($x,y,z$) are the true position of the object from which these points are reflected. Thus we can directly discretize the coordinates to calculate occupancy features for each BEV cell.
For points in each sweep, the $x$ and $y$ dimensions are discretized to create a BEV grid and the $z$ dimension is discretized to create a multi channel binary occupancy feature for each cell in the grid. These temporal, per sweep occupancy features are concatenated and used as input to a lightweight network to learn spatio-temporal lidar features for the BEV cells.
Additional details about feature learning from lidar and HD map data can be found in the supplementary material.

\subsection{Backbone Network Architecture}
\vspace{-0.8em}
We use the same backbone network architecture as~\cite{pnpnet, multixnet}. In particular, we use the multi-scale inception module~\cite{pnpnet} as the main building block for extracting and combining multiple scale features. The inception block contains three branches, each with a down-sampling ratio of 1x, 2x and 4x. We use 3 sequential inception blocks and then use a feature pyramid network to compute the features for each BEV cell. We concatenate features from all modalities at multiple scales and provide it as input to the first inception block. The multiscale features for each modality are computed using a single strided convolution. 
A dense single stage convolutional header~\cite{pixor} is used for detecting objects from the pyramid features. For trajectory prediction, we extract a large rotated ROI~\cite{spagnn, multixnet} of $40m$x$40m$ centered at the object to learn actor centric features which are then used to predict the final trajectory~\cite{multixnet,mtp,multipath}. 

\subsection{End-to-End Training}
\vspace{-0.8em}
We jointly train the proposed method using a multi-task loss defined as a weighted sum of the detection and trajectory loss:  $\mathcal{L}_{total} = \mathcal{L}_{det} + \lambda \mathcal{L}_{traj}$.

\textbf{Detection Loss ($\mathcal{L}_{det}$)} is a multi-task loss defined as a weighted sum of classification and regression loss: $\mathcal{L}_{det} = \mathcal{L}^{cls}_{det} + \alpha \mathcal{L}^{reg}_{det}$. We use focal loss~\cite{retinanet} to train classification of each BEV cell for being at the center of an object class. We parameterize each object $i$ by it's center $(x_{i}, y_{i})$, orientation ($\theta_{i}$) and size ($w_{i}, h_{i}$). We use smooth L1 loss to train regression parameters of each object. 

\textbf{Trajectory Loss ($\mathcal{L}_{traj}$):} To evaluate the generality of our proposed method, we consider multiple different trajectory formulations: single-hypothesis~\cite{multixnet} and multi-hypothesis~\cite{mtp, multipath}. We consider each waypoint at time $t$ of a trajectory $j$ to be a 2D Laplace distribution parameterized by its position ($x_{j}^{t}, y_{j}^{t}$) and scale ($\sigma_{j,x}^{t}, \sigma_{j,y}^{t}$). We use the sum of KL divergence~\cite{lasernet-kl} between the ground truth and predicted distribution for all the waypoints as our regression loss for trajectory. In case of single hypothesis prediction, the prediction loss only contains the regression component, $\mathcal{L}_{traj}$ = $\mathcal{L}^{reg}_{traj}$. However, in multiple-hypothesis prediction, it contains both regression and classification, $\mathcal{L}_{traj}$ = $\mathcal{L}^{reg}_{traj} + \beta \mathcal{L}^{cls}_{traj}$. We use cross-entropy loss for learning the confidence of each predicted hypothesis.  

\section{Experimental Results}
\label{sec:result}
\textbf{Datasets:} We evaluate our approach on two large-scale autonomous driving datasets: nuScenses and \datasetname. nuScense is publicly available and {\datasetname} is collected in-house in dense urban settings. For each lidar sweep in these datasets, we find the nearest (in time) radar sweep to synchronize them. {\datasetname} contains an order of magnitude more data than nuScenes with $\sim$2.5 million sweeps for training and $\sim$600k for validation.

For nuScenes, we use the official ROI of $100m$ square centered at the SDV. Whereas, for {\datasetname} dataset we use an ROI of square of $100m$ in front of the vehicle. For training and testing the trajectory prediction we use the sensor data from past 0.5 seconds and predict the trajectory for 3 seconds in future. We use objects of vehicle class for evaluation in both datasets. For additional training details, refer to the supplementary material.

\textbf{Metrics:} For evaluating trajectory predictions for each object we use: Average Displacement Error (\text{ADE}) and Final Displacement Error (\text{FDE}) defined as following:

\vspace{-1.2em}
\begin{equation}
    \text{ADE} = \frac{1}{4}\sum_{t = 0}^{3} ||\bm{x}_{t} - \bm{\hat{x}}_{t}||_{2} \,\,\,\,\,\,\,\,\, \text{FDE} = ||\bm{x}_{3} - \bm{\hat{x}}_{3}||_{2}
\end{equation}
\vspace{-1.2em}

where $\bm{\hat{x}}_{t}$ is the predicted position at time $t$ (in s) and $\bm{x}_{t}$ is the ground truth position. To calculate these metrics, we consider a detection to be true positive if it has an IoU $\ge 0.5$ with the ground truth. We use the same recall value of 80\% for comparing methods with different detection sets. We train and test on the vehicle class in both datasets.

\begin{table}[t]
    \centering
    \tablestyle{5pt}{1.08}
    \begin{tabular}{l|c|cc|cc|cc|cc}
        \multirow{3}{*}{Method} & \multirow{3}{*}{Radar} & \multicolumn{4}{c|}{60\% Recall} & \multicolumn{4}{c}{80\% Recall}\\ \cline{3-10}
        & &\multicolumn{2}{c|}{All} & \multicolumn{2}{c|}{Moving} & \multicolumn{2}{c|}{All} & \multicolumn{2}{c}{Moving} \\ \cline{3-10}
        & & $\text{ADE}\downarrow$ & $\text{FDE}\downarrow$ & $\text{ADE}\downarrow$ & $\text{FDE}\downarrow$ & $\text{ADE}\downarrow$ & $\text{FDE}\downarrow$ & $\text{ADE}\downarrow$ & $\text{FDE}\downarrow$
         \\ \shline
        SpAGNN \cite{spagnn} & $\times$        & - & 145 & - & - & - & - & - & - \\
        LaserFlow \cite{laserflow} & $\times$  & - & 153 & - & - & - & - & - & - \\
        RV-FuseNet \cite{rvfusenet} & $\times$ & - & 123 & - & - & - & - & - & - \\
        MultiXNet* \cite{multixnet}& $\times$ & 62 & 113 & 141 & 287 & 68 & 122 & 157  & 314 \\
        {\networkname} (Ours) & \checkmark           & \textbf{56} & \textbf{102} & \textbf{127} & \textbf{263} & \textbf{63} & \textbf{106} & \textbf{136}  & \textbf{278} \\
    \end{tabular}
    \vspace{0.5em}
    \caption{\textbf{Trajectory prediction performance on nuScenes\cite{nuscenes}}.
    Both $\text{ADE}$ and $\text{FDE}$ are in cm.
    \vspace{-1.7em}
    }
    \label{tab:stoa-nuscenes}
\end{table}

\begin{table}[t]
    \centering
    \tablestyle{5pt}{1.1}
    \begin{tabular}{l|c|cc|cc}
        \multirow{2}{*}{Method} & \multirow{2}{*}{Radar} & \multicolumn{2}{c|}{All} & \multicolumn{2}{c}{Moving}\\ \cline{3-6}
        & & $\text{ADE}\downarrow$ & $\text{FDE}\downarrow$ & $\text{ADE}\downarrow$ & $\text{FDE}\downarrow$
         \\ \shline
        MultiXNet*~\cite{multixnet} & $\times$ &         \xdcolortext{35} & \xdcolortext{56} & \xdcolortext{154} & \xdcolortext{326} \\
        {\networkname} (Ours) & \checkmark             & \textbf{\xdcolortext{33}} & \textbf{\xdcolortext{52}} & \textbf{\xdcolortext{135}} & \textbf{\xdcolortext{285}}\\
    \end{tabular}
    \vspace{0.5em}
    \caption{
    \textbf{Trajectory prediction performance on \xdcolortext{{\datasetname}}}. 
    Both $\text{ADE}$ and $\text{FDE}$ are in cm.
    \label{tab:stoa-x17k}
    }
    \vspace{-1.8em}
\end{table}

\subsection{Comparison with State of the Art} 
\vspace{-0.8em}

\textbf{Single Hypothesis Methods:} Table~\ref{tab:stoa-nuscenes} shows the comparison of our approach with previous single-hypothesis approaches on nuScenes. Note that here we additionally report prediction results on 60\% recall points since all existing method only report results in this setting and for fair comparison we compare our method in their setting. 
We use a state-of-the-art model MultiXNet~\cite{multixnet} as our lidar and map based baseline and show that adding radar significantly improves results on nuScenes~\ref{tab:stoa-nuscenes}.

In particular we get a substantial  $\sim13\%$ reduction in the prediction error at 3s for moving objects. This demonstrates that the proposed approach of learning feature is effectively able to exploit radar velocity information. Furthermore, using radar we achieve state-of-art results on this dataset.

Table~\ref{tab:stoa-x17k} demonstrates the impact of incorporating radar into MultiXNet~\cite{multixnet} on our in-house large-scale dataset ({\datasetname}). Here we see a similar $\sim13\%$ reduction in prediction error for moving objects. This dataset is collected in dense urban settings which inherently contain a lot of static vehicles ($\sim85\%$ of all the labels). Overall improvement on all vehicles is therefore lower since the presence of a relatively high proportion of static vehicles masks the improvement on the moving vehicles.

\begin{figure}[t]
    \centering
    \includegraphics[width=\textwidth]{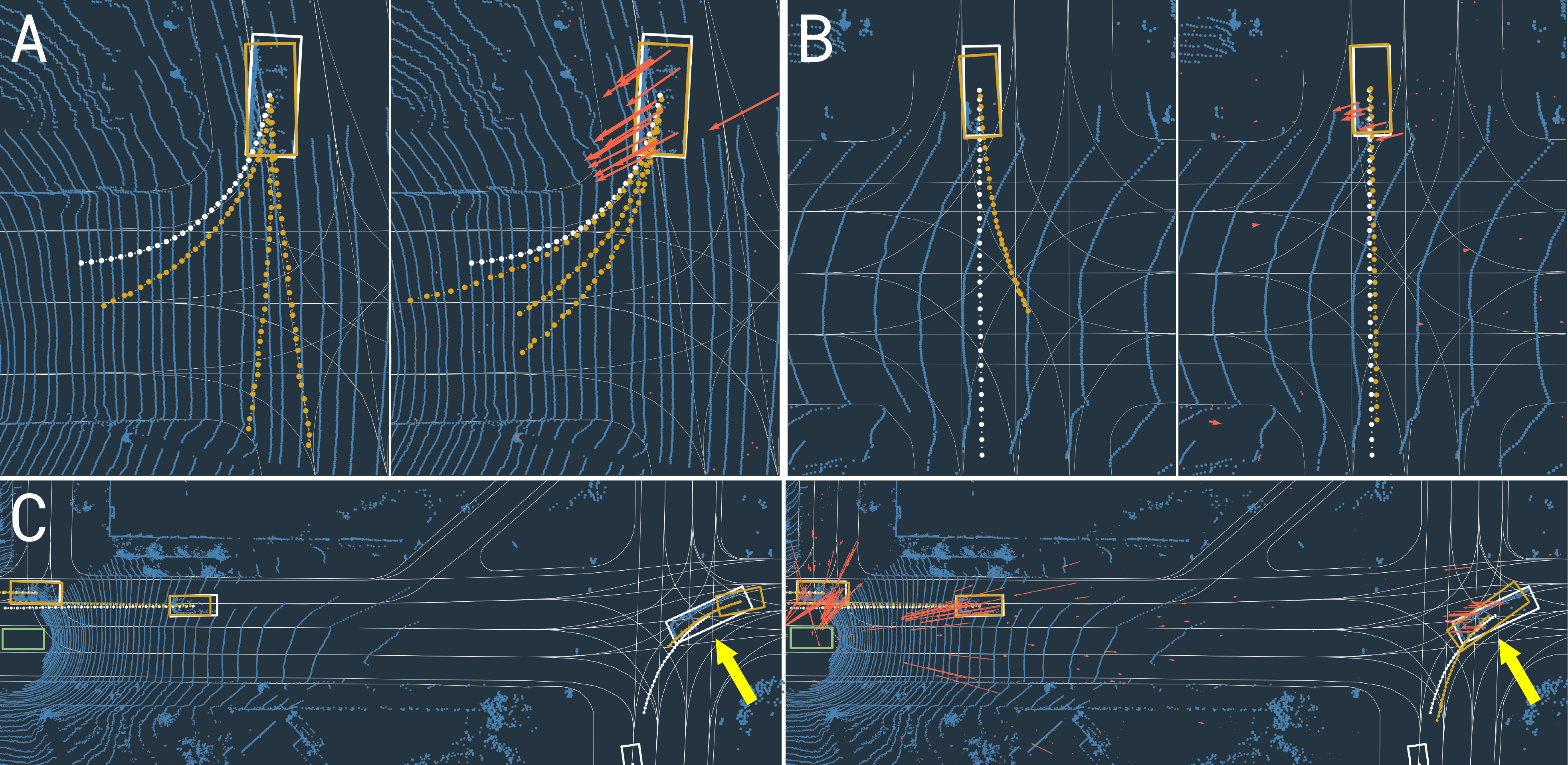}
    \caption{
        \textbf{Qualitative Improvements:}
        These three example scenes from {\datasetname} show improvements with the addition of radar data. Labels are shown in white and predicted output is shown in \predcolortext{gold}. 
        In \textbf{(A)}, the vehicle just started slowing down (has high deceleration) for the turn. This change in velocity can be hard to measure with historical lidar alone so the lidar model still gives a high probability for the mode which goes straight. With radar however, all of the modes correctly indicate that the vehicle is turning.
        \textbf{(B)} shows a case where lidar is sparse due to occlusion which leads to an incorrect estimation of the object turning. Since radar shows approximately zero radial velocity, with radar the model can correctly infer that it is going straight at high velocity.
        In \textbf{(C)}, the vehicle pointed at by the yellow arrow is too far away the lidar model to correctly estimate the shape, heading or velocity. However, with radar radial velocity observations all can be correctly estimated.
    \vspace{-0.8em}
    }
    \label{fig:qualitative_improvements}
\end{figure}

\textbf{Multiple Hypothesis Methods:} We incorporate anchorless MTP~\cite{mtp} and anchor-based MultiPath~\cite{multipath} multi-hypothesis prediction formulations in our proposed network and study the impact of adding radar. We use 16 hypotheses for both MTP and MultiPath. Table~\ref{tab:multi-hypothesis} shows the results on moving objects. We can see that adding radar data helps on both formulations and in both datasets. We also see that the performance improvement is higher when we chose a lower number of hypotheses. This clearly shows that adding radar is not only able to produce better hypotheses but is able to do so with higher confidence as can be seen in Fig.~\ref{fig:qualitative_improvements}.

\begin{table}[t]
    \vspace{-1.5em}
    \tablestyle{5pt}{1.1}
    \centering
    \begin{tabular}{l|c|ccccc}
        Method & Radar & $\text{ADE}_{1}\downarrow$ & $\text{ADE}_{3}\downarrow$ & $\text{ADE}_{5}\downarrow$ & $\text{ADE}_{10}\downarrow$ &  $\text{ADE}_{15}\downarrow$
         \\ \shline
        MTP~\cite{mtp} & $\times$              & 162 \text{\textbar} \xdcolortext{184} & 142 \text{\textbar} \xdcolortext{137} & 122 \text{\textbar} \xdcolortext{116} & 94 \text{\textbar} \xdcolortext{\textbf{86}}  & 76 \text{\textbar} \xdcolortext{\textbf{77}}  \\
        MTP(Ours) &  \checkmark & \textbf{150} \text{\textbar} \xdcolortext{\textbf{181}} & \textbf{112} \text{\textbar} \xdcolortext{\textbf{132}} & \phantom{0}\textbf{95}  \text{\textbar} \xdcolortext{\textbf{112}} & \textbf{70} \text{\textbar} \xdcolortext{\textbf{86}}  & \textbf{65} \text{\textbar} \xdcolortext{\textbf{77}}\\
        \hline
        MultiPath~\cite{multipath} & $\times$  & 206 \text{\textbar} \xdcolortext{180} & 122 \text{\textbar} \xdcolortext{108} & 102 \text{\textbar} \xdcolortext{\phantom{0}95}  & 96 \text{\textbar} \xdcolortext{90}  & 76 \text{\textbar} \xdcolortext{89}\\
        MultiPath(Ours) & \checkmark           & \textbf{168} \text{\textbar} \xdcolortext{\textbf{163}} & \phantom{0}\textbf{95} \text{\textbar} \xdcolortext{\textbf{101}} & \phantom{0}\textbf{81}  \text{\textbar} \xdcolortext{\phantom{0}\textbf{91}}  & \textbf{75} \text{\textbar} \xdcolortext{\textbf{87}}  & \textbf{\textbf{74}} \text{\textbar} \xdcolortext{\textbf{85}}\\
    \end{tabular}
    \vspace{0.5em}
    \caption{
    \textbf{Trajectory prediction performance on \textit{moving} objects using multi-hypothesis formulations}. 
    $\text{ADE}_{k}$ (in cm) denotes the minimum of the $\text{ADE}$ over top k hypothesis on nuScenes (black) and \xdcolortext{\datasetname (blue)}
    }
    \label{tab:multi-hypothesis}
\end{table}

\subsection{Comparison on Different Object Characteristics}
\vspace{-0.8em}

\begin{figure}[t]
    \centering
    \includegraphics[width=\textwidth]{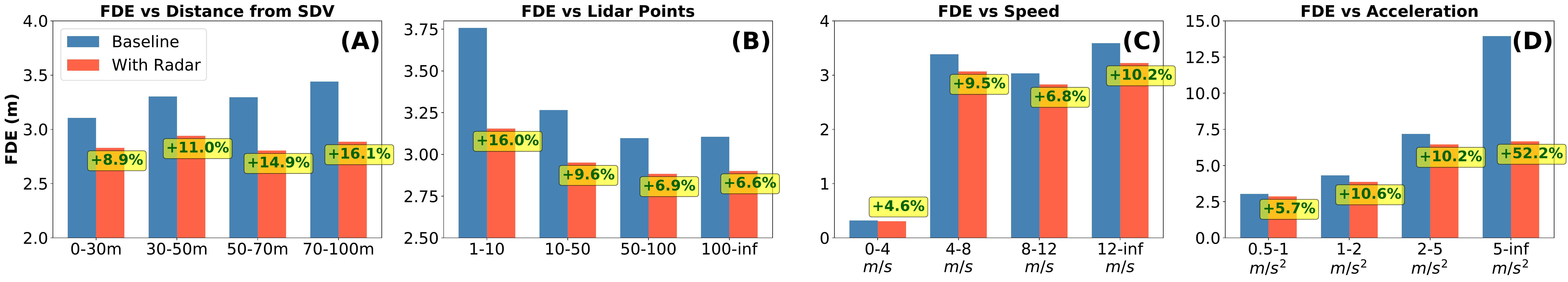}
    \vspace{-1.5em}
    \caption{
    \textbf{Results by different object characteristics:}
    We show relative improvement (\%) on {\datasetname} in $\text{FDE}$ metric (m) by adding radar. 
    ~\textbf{(A)} is binned by distance to AV, 
    ~\textbf{(B)} is binned by number of lidar points,
    ~\textbf{(C)} is binned by speed,
    ~\textbf{(D)} is binned by acceleration (the absolute value).
    We can infer that adding radar significantly improves over baselines in several challenging scenarios.
    \vspace{-1.8em}
    }
    \label{fig:ablation/ablation_x17k}
\end{figure}

Radar provides improvements in cases where lidar has difficulty in learning object dynamics. 
Figure~\ref{fig:ablation/ablation_x17k} shows the improvements in $\text{FDE}$ of objects over different object characteristics on {\datasetname} in the single hypothesis model. In the following, we analyze some of the factors where trajectory prediction is particularly challenging.

\textbf{Position and Scene based Factors:} We first look at objects based on the number of lidar points on them and their distance from the SDV. In Figure~\ref{fig:ablation/ablation_x17k}(b), we see significant gains in prediction performance as the number of lidar points on an object decreases. Implicit association of sparse point clouds across time for learning the dynamics is a difficult problem. Therefore, having a direct measurement of velocity reduces the error by $\sim16\%$ in the case of objects with as few as 1-10 lidar points. Similarly, since distance from SDV is inversely related to the number of lidar points, Figure~\ref{fig:ablation/ablation_x17k}(a) shows that the improvements from radar increase at longer ranges.

\textbf{Object Dynamics based Factors:} We now look at objects based on their speed and acceleration. Figure~\ref{fig:ablation/ablation_x17k}(c) shows that radar can improve predictions on objects regardless of speed. The very low speed ($\le 4m/s$) objects are usually easier to associate across time, so the positional information of lidar alone is usually enough for learning dynamics. Figure~\ref{fig:ablation/ablation_x17k}(d) shows the improvements of radar based on acceleration of the object. We can clearly see that radar improves predictions by a large margin as the acceleration increases. Since it takes more observations of lidar (positional information) to learn acceleration as compared to radar (velocity information), it reduces the error by more than 50\% for objects with very high acceleration of greater than $5$m$/s^{2}$. 

\begin{table}[t]
    \vspace{-0.8em}
    \tablestyle{5pt}{1.1}
    \centering
    \subfloat[][Impact of Graph Parameters] {
    \begin{tabular}{l|c|cc}
          K & d & $\text{ADE}\downarrow$ & $\text{FDE}\downarrow$
          \\ \shline
         1 & 1      & 140 \text{\textbar} \xdcolortext{142} & 295 \text{\textbar} \xdcolortext{310}\\
         1 & 10     & \textbf{136} \text{\textbar} \xdcolortext{\textbf{135}} & \textbf{278} \text{\textbar} \xdcolortext{\textbf{285}}\\
         1 & 25     & 137 \text{\textbar} \xdcolortext{138} & 279 \text{\textbar} \xdcolortext{302}\\
         1 & +Inf   & 140 \text{\textbar} \xdcolortext{140} & 284 \text{\textbar} \xdcolortext{301}\\
         2 & +Inf   & 142 \text{\textbar} \xdcolortext{145} & 290 \text{\textbar} \xdcolortext{315}\\
     \end{tabular}
     \label{tab:ablation_graph}
     }
    \subfloat[][Impact of Grid Resolution] {
    \begin{tabular}{l|c|cc}
         Res. & $\text{ADE}\downarrow$ & $\text{FDE}\downarrow$
          \\ \shline
         0.125x & \textbf{136} \text{\textbar} \xdcolortext{143} & \textbf{280} \text{\textbar} \xdcolortext{311}\\
         0.25x  & 140 \text{\textbar} \xdcolortext{\textbf{142}} & 295 \text{\textbar} \xdcolortext{\textbf{310}}\\
         0.5x   & 141 \text{\textbar} \xdcolortext{146} & 291 \text{\textbar} \xdcolortext{316}\\
         1x     & 146 \text{\textbar} \xdcolortext{150} & 297 \text{\textbar} \xdcolortext{324}\\
     \end{tabular}
     \vspace{1.25em}
     \label{tab:ablation_resolution}
     }
    \subfloat[][Impact of Radar History] {
    \begin{tabular}{l|c|cc}
         History & $\text{ADE}\downarrow$ & $\text{FDE}\downarrow$
          \\ \shline
          $0s$ & 157 \text{\textbar} \xdcolortext{154} & 314 \text{\textbar} \xdcolortext{326} \\
         $0.1s$ & 138 \text{\textbar} \xdcolortext{142} & 285 \text{\textbar} \xdcolortext{307} \\
         $0.2s$ & \textbf{136} \text{\textbar} \xdcolortext{141} & 281 \text{\textbar} \xdcolortext{310}\\
         $0.5s$ & \textbf{136} \text{\textbar} \xdcolortext{\textbf{135}} & \textbf{278} \text{\textbar} \xdcolortext{\textbf{285}}\\
     \end{tabular}
     \vspace{1.25em}
    \label{tab:ablation_history}
     }
     \caption{
     \textbf{Ablation studies:}
     We study the impact of different parameters of our feature learning scheme for \textit{moving} objects on nuScenes and \xdcolortext{\datasetname} for the single hypothesis setting.
     $K$ and $d$ denote the number of neighbors and the distance threshold (in number of cells) respectively. \textit{Res} denotes the resolution as ratio of the size of the radar feature extraction grid to the lidar occupancy grid. History denotes the size of temporal context for radar. 
     Using this study, we chose $0.25$x resolution, $K=1$, $d=10$ and a radar history of $0.5s$ as our final model.
     \vspace{-1.8em}
     }
\end{table}

\subsection{Ablation of Radar Feature Learning}
\vspace{-0.8em}

We have proposed multiple design strategies to address the challenges for fusion of radar with other sensors. In the following section, we analyze the importance of those design choices. 

\textbf{Graph Parameters:} Table~\ref{tab:ablation_graph} shows the impact of various parameters of the graph for association of BEV cells with radar points. From the table we can see that using a reasonably wide area to find the neighbors can improve the results significantly by $\sim8\%$ as compared to sparse association ($K=1$ and $d=1$). These results clearly show that for achieving good performance, it is necessary to extract features for BEV cells with points which do not directly fall in them. Additionally, we observe that having higher number of neighbors does not provide additional value due to the sparsity. 

\textbf{BEV Resolution:} Table~\ref{tab:ablation_resolution} shows the impact of BEV cell resolution for learning radar based features. Due to sparsity of radar we can see that selecting a resolution that is $0.25$ times lower than lidar can significantly improve the impact of adding radar on both datasets. 

\textbf{Radar History:} Table~\ref{tab:ablation_history} shows the impact of temporal radar context. Since radar only provides radial component of velocity, multiple observations on an object are needed to implicitly learn the tangential component. For objects with a high number of radar points, it may be possible to learn this from a single sweep. On objects with sparse radar points, however, these observations have to come from past frames. This clearly shows value from increasing temporal context for radar. The performance on {\datasetname} improves more than nuScenes due to the longer average range of objects in {\datasetname} where the sparsity of radar is more of an issue.    

\vspace{-0.8em}
\section{Conclusion}
\label{sec:conclusion}
\vspace{-0.8em}
In this work we proposed using instantaneous velocity information of radar for improving the end-to-end trajectory prediction. 
To this end, we proposed an efficient spatio-temporal radar feature extraction scheme which achieves state-of-the-art performance on multiple large-scale datasets.
We further demonstrated the advantages leveraging rich dynamic information from radar in challenging scenarios where lidar struggles. 
Our approach can increase overall robustness of end-to-end trajectory prediction systems, particularly in scenarios with rapidly changing dynamics where lidar-only baselines struggle and potentially reduce harmful interactions.

\clearpage


\makeatletter
\renewcommand{\@maketitle}{
\newpage
 \null
 \vskip 2em%
 \begin{center}%
  {\LARGE\bf \@title \par}%
 \end{center}%
 \par
 \vskip 0.3in \@minus 0.1in
 } \makeatother

\title{Appendix}
\author{}
\maketitle

In this appendix, we first outline implementation details for training LiRaNet. Then, we discuss the network architecture in detail. Next, we discuss additional results in terms of detection performance, and analysis based on different object characteristics. We then analyze the run-time impact of adding radar and present additional qualitative results.

\section{Implementation Details}
We first describe the implementation details of our proposed method, LiRaNet, on both datasets. On nuScenes, we discretize $x$ and $y$ axis with a resolution of 10 cells per meter and $z$ axis with a resolution of 5 cells per meter. We use data augmentation to further increase the amount of training data on nuScenes. We interpolate labels from key frames to non-key frames to increase the number of training samples. We further increase the training data by randomly rotating, translating and scaling point clouds and labels. All past point clouds and future trajectory labels are augmented using the same random transformation. For each training sample, we use translation between $[-1, 1]m$ for the $x$ and $y$ axes, and $[-0.2, 0.2]m$ for the $z$ axis. We use a rotation between $[-\sfrac{\pi}{4}, \sfrac{\pi}{4}]$ about the $z$ axis and a scaling between $[0.95, 1.05]$ for all 3 axes.

On X17k, we discretize the $x$ and $y$ axis with a resolution of 6.4 cells per meter and $z$ axis with a resolution of 5 cells per meter. This dataset contains labels for all the frames and has an order of magnitude more data than nuScenes. Thus, we do not perform any data augmentation. 

Similar to previous work~\cite{laserflow, rvfusenet}, on nuScenes, we define our \textit{vehicle} class to contain car, truck, bus, trailer and construction vehicles. On X17k, the same definition of the vehicle class is used. For both datasets, we only train and test on the objects which contain at least one lidar point. During training, we define a BEV cell as positive if it contains the center of the object. We use Adam optimizer and train with batch size of 32 on 16 GPUs for 2 epochs. We use 0.0004 as initial learning rate and decay it by 0.1 after 1.5 and 1.9 epochs. 

\section{Network Architecture and Object Detection}
\begin{figure}[ht]
    \centering
    \includegraphics[width=\textwidth]{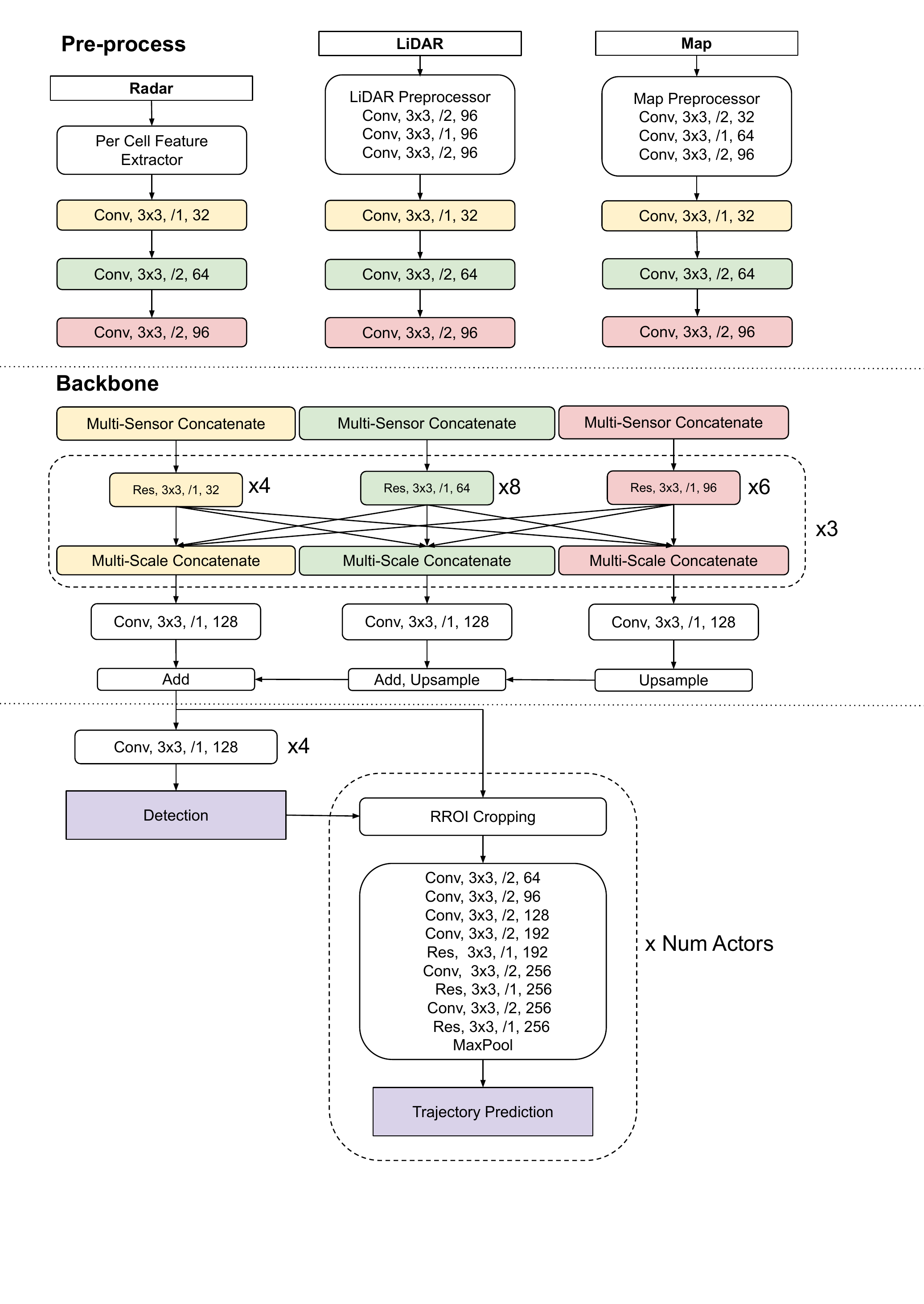}
    \caption{
    \textbf{Network Architecture:} This figure show the detailed architecture of LiRaNet. Radar is fused with LiDAR and Map to generate multi-sensor feature volume in BEV which is used for detection and trajectory prediction. Each learnable operator is denoted as (\textit{op, kernel, stride, number of features}). \textit{Conv} denotes convolution and \textit{Res} denotes a residual block.
    }
    \label{fig:network}
\end{figure}

\vspace{-1em}
Figure~\ref{fig:network} shows our network architecture for end-to-end trajectory prediction using lidar, radar and HD maps. Each sensor is pre-processed to generate multi-scale features. These features for all scales are concatenated and further processed using a backbone CNN to extract and combine multi-scale features using all the sensors. The output features of the backbone are used for detection using a full convolutional single shot detector. For each cell we predict a class probability and a bounding box using the multi-sensor features. We then pick top 200 cells using the predicted class probability and do NMS to remove duplicates. For each of the remaining detections (i.e. objects) we predict the trajectory using a crop of the same feature map used for detection.For each detected object, a large 40m x 40m rotated ROI centered around the object is cropped from the output feature volume produced by the backbone network. This crop is further passed through a CNN to generate final trajectory predictions.

\begin{figure}[!ht]
    \centering
    \includegraphics[width=\textwidth]{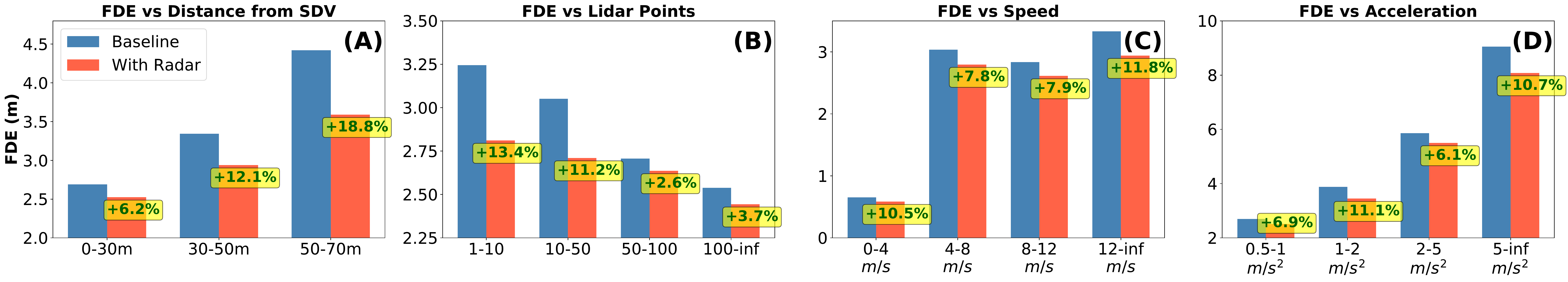}
    \vspace{-1.5em}
    \caption{
    \textbf{Results by different object characteristics:}
    We show relative improvement (\%) on nuScenes in $\text{FDE}$ metric (m) by adding radar. 
    ~\textbf{(A)} is binned by distance to AV, 
    ~\textbf{(B)} is binned by number of lidar points,
    ~\textbf{(C)} is binned by the speed of the object,
    ~\textbf{(D)} is binned by the absolute value of acceleration of the object.
   These demonstrate that adding radar significantly improves over baselines in several challenging scenarios.
    \vspace{-1.5em}
    }
    \label{fig:ablation/ablation_nuscenes}
\end{figure}

\section{Comparison on Different Object Characteristics for nuScenes}
Figure~\ref{fig:ablation/ablation_nuscenes} shows improvement in trajectory prediction by adding radar along different factors which make trajectory prediction particularly challenging. Similar to X17k, we can observe that radar helps in all cases. It is especially beneficial in cases which are more challenging for lidar such as low lidar point density, rapidly changing dynamics due to acceleration, etc.

\section{Detection Performance}
In addition to the marked improvement on trajectory prediction as shown in the main paper, we have observed significant improvements in detection performance.  We evaluate detection performance using the standard AP metric using IoU based association between ground truth and predicted bounding boxes. As we can observe from Table~\ref{tab:detection}, addition of radar improves the detection performance on both datasets. We observe larger gains on nuScenes relative to {\datasetname} due to the relatively high sparsity of lidar points in nuScenes.

\begin{table}[!ht]
    \tablestyle{5pt}{1.08}
    \centering
    \begin{tabular}{l|c|cc|cc}
        \multirow{2}{*}{Method} & \multirow{2}{*}{Radar} & \multicolumn{2}{c|}{nuScenes} & \multicolumn{2}{c}{\xdcolortext{X17k}}\\ \cline{3-6}
        & & $\text{AP}_{0.5}\uparrow$ & $\text{AP}_{0.7}\uparrow$ & $\text{AP}_{0.5}\uparrow$ & $\text{AP}_{0.7}\uparrow$
         \\ \shline
        MultiXNet*~\cite{multixnet} & $\times$ & 71.54 &  58.47 & \xdcolortext{84.02} & \xdcolortext{74.13} \\
        {\networkname} (Ours) & \checkmark     & \textbf{79.11} & \textbf{63.67} & \textbf{\xdcolortext{86.32}} & \textbf{\xdcolortext{76.02}}\\
    \end{tabular}
    \vspace{0.5em}
    \caption{
    \textbf{Improvement in detection performance by adding radar}.
    For nuScenes, we define our \textit{vehicle} class to contain the car, truck, bus, trailer and construction vehicles. On \xdcolortext{\datasetname}, the same definition of the vehicle class is used.
    \label{tab:detection}
    }
    \vspace{-0.5em}
\end{table}

\begin{table}[!ht]
    \tablestyle{5pt}{1.08}
    \centering
    \begin{tabular}{l|c|c|c}
        Method & Radar & Radar History (s) & Latency (ms)
        \\ \shline
        MultiXNet*~\cite{multixnet} & $\times$ &   -    &  38.02\\
        {\networkname} (Ours) & \checkmark     & $0.3s$ &  38.51\\
        {\networkname} (Ours) & \checkmark     & $0.4s$ &  39.96\\
        {\networkname} (Ours) & \checkmark     & $0.5s$ &  40.98\\
    \end{tabular}
    \vspace{0.5em}
    \caption{
    \textbf{Runtime Analysis on \xdcolortext{\datasetname}}. All results are on a GeForce RTX 2080 Ti.
    \label{tab:runtime}
    }
    \vspace{0.5em}
\end{table}

\vspace{-2.5em}
\section{Runtime Analysis}
Runtime latency is of prime importance for end-to-end trajectory prediction systems in order to deploy them in realtime applications.

The primary components of {\networkname} impacting runtime performance (relative to the lidar-only baselines) are: 1) the per cell feature extraction module and 2) the temporal fusion layers.
We implement the per-cell feature extraction (including the k-nearest neighbour search) module as optimized CUDA~\cite{cuda} kernels.
Since per cell feature extraction from radar points doesn't depend on data from other sensors, we pipeline it alongside with the map and lidar preprocessing modules to further reduce effective latency.
The temporal fusion layers along with the core network backbone are further optimized for inference using TensorRT~\cite{tensort}.
All results are computed on a GeForce RTX 2080 Ti GPU using batch size 1 and are averaged over 500 iterations.
Latency of our network along with the baselines can be found in Table~\ref{tab:runtime}. 
As we can see, overall impact of adding radar on runtime is relatively limited, $< 3ms$, even for the case of longer time horizon, $0.5 s$ of radar history. 

\section{Qualitative Results}
\begin{figure}[ht]
    \centering
    \includegraphics[width=\textwidth]{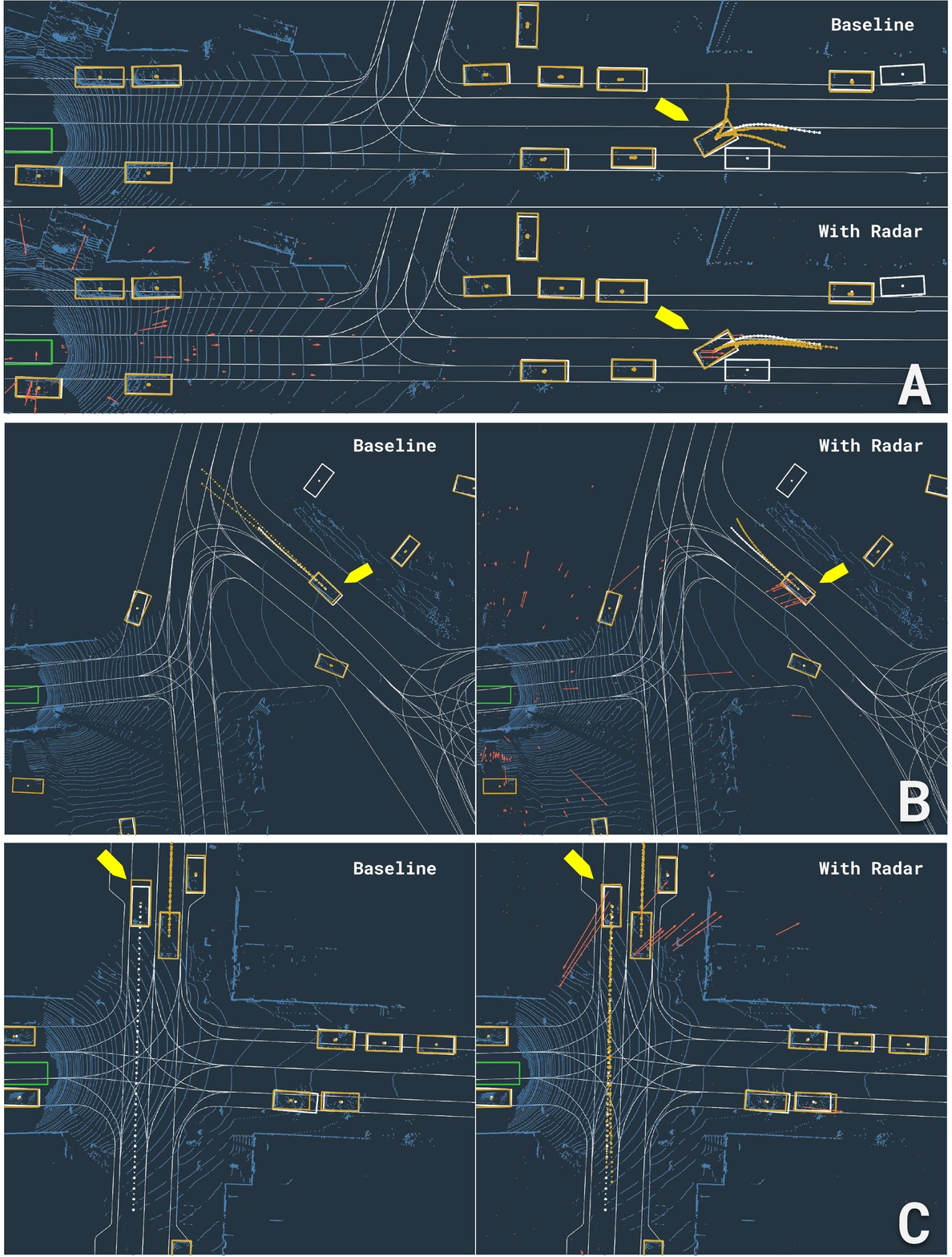}
    \caption{
    \textbf{Improvements in Prediction.}
    These three scenes show where radar improves trajectory prediction. 
    The vehicle in \textbf{A} is accelerating into the street. Given the range and high acceleration of the vehicle, radar shows a clear benefit in estimating object trajectories with respect to the lidar only baseline.
    Similarly, \textbf{B} shows a case where the addition of radar helps accurately predict the motion of a vehicle slowing down for a turn.
    In \textbf{C} a vehicle has just come from an area of occlusion resulting in sparse historical lidar data. The radar data lets the model quickly estimate an accurate future trajectory.
    }
    \label{fig:pred}
\end{figure}

\begin{figure}[ht]
    \centering
    \includegraphics[width=\textwidth]{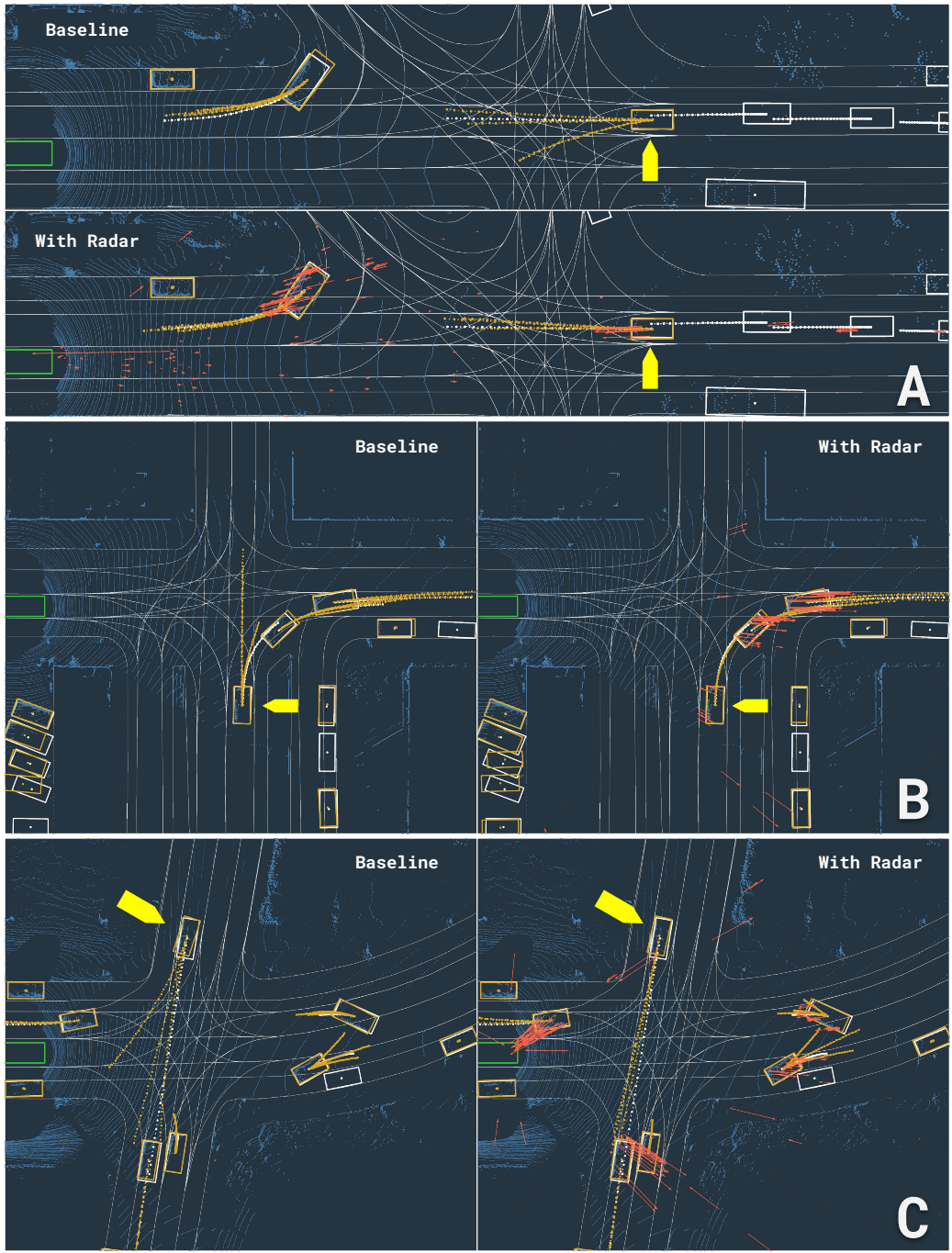}
    \caption{
    \textbf{Improvements in Mode Ambiguity Resolution.}
    Here we show three cases where radar helps resolve ambiguities in future trajectory prediction. We show the top 3 hypotheses.
    Without radar, \textbf{A} predicts a hypothesis that the vehicle can turn left in front of the SDV with high confidence.
    Without radar, the vehicle in \textbf{B} is predicted to cross SDV's path. Mispredictions like these could lead to uncomfortable hard breaking.
    In \textbf{C}, the vehicle trajectory predictions converge when the velocity information from (even sparse) radar is incorporated.
    In all cases addition of radar leads to more accurate mode prediction.
    }
    \label{fig:intent}
\end{figure}

\begin{figure}[ht]
    \centering
    \includegraphics[width=\textwidth]{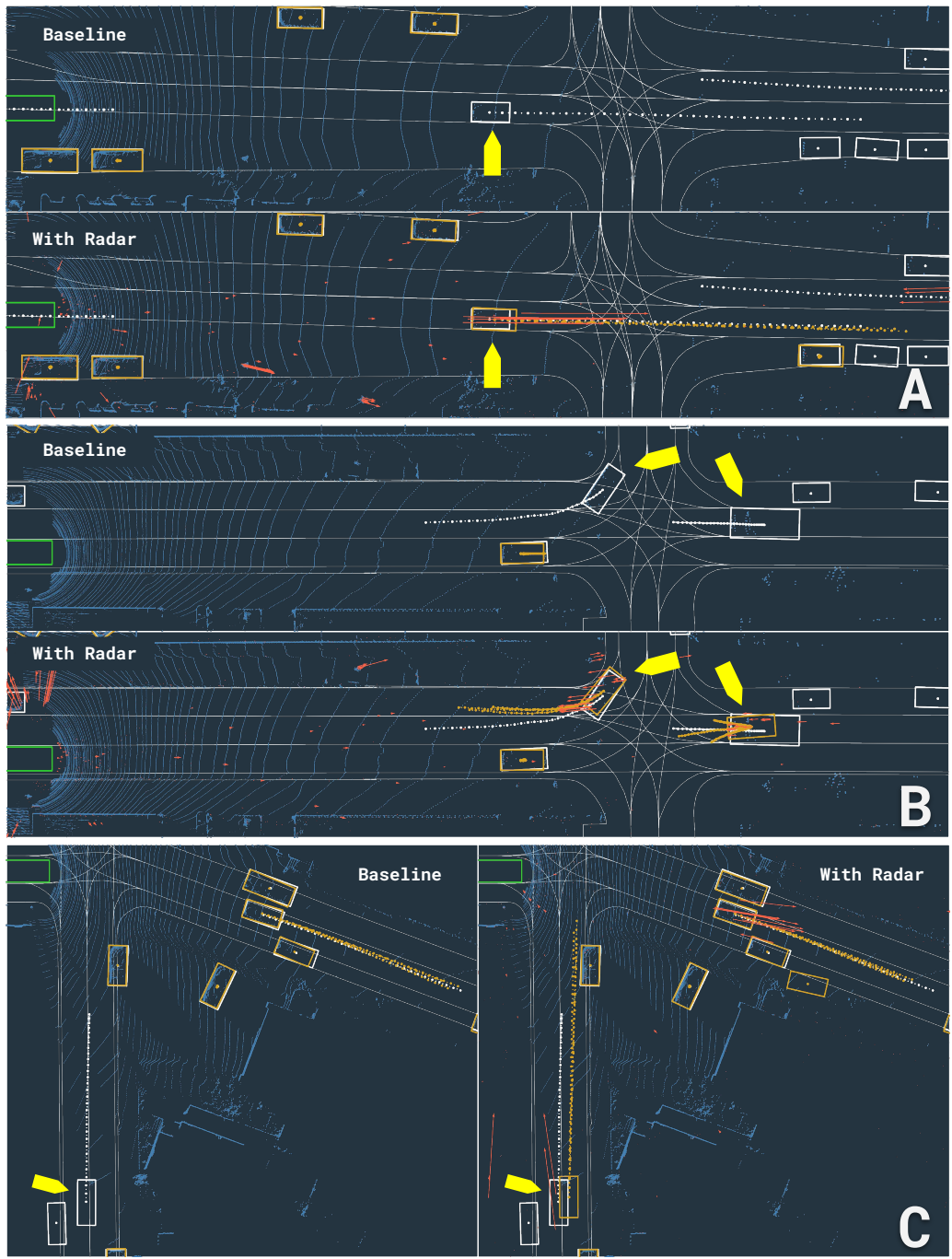}
    \caption{
    \textbf{Improvements in Detection.}
    These three scenes show where addition of radar improves the ability to detect and consequently predict trajectories of objects.
    In scene \textbf{A} the incline of the road and range of the object limit the number of lidar returns on the object. This results in less confident detections without radar.
    The two vehicles in \textbf{B} have been occluded by other objects in the scene. These objects predominantly detected by radar are noisier than if lidar was present.
    Similarly, in \textbf{C} the vehicle is emerging from occlusion. The number of lidar points on the object is too low to produce a detection without radar.
    }
    \label{fig:det}
\end{figure}

Figures~\ref{fig:pred}, ~\ref{fig:intent} and ~\ref{fig:det} show scenes from X17k where adding radar helps produce better trajectory predictions. We show \lidarcolortext{lidar points with light blue}, \radarcolortext{radar point velocities with orange}, labels with white and \predcolortext{model predictions with gold}. The sensor data used in the model is depicted in the scene.

\clearpage

\bibliography{main}  
\end{document}